\documentclass[letterpaper]{article} 
\usepackage[submission]{aaai25}  
\usepackage{times}  
\usepackage{helvet}  
\usepackage{courier}  
\usepackage[hyphens]{url}  
\usepackage{graphicx} 
\usepackage{natbib}  
\usepackage{caption} 
\usepackage{algorithm}
\usepackage{algorithmic}

\usepackage{amsmath}
\usepackage{amsfonts}
\usepackage{multirow}
\usepackage{makecell}
\usepackage{xcolor,colortbl}
\usepackage{color}
\usepackage{booktabs}

\usepackage{newfloat}
\usepackage{listings}
\DeclareCaptionStyle{ruled}{labelfont=normalfont,labelsep=colon,strut=off} 
\floatstyle{ruled}
\newfloat{listing}{tb}{lst}{}
\floatname{listing}{Listing}
\title{MePT: Multi-Representation Guided Prompt Tuning for Vision-Language Model}
\author {
    Xinyang Wang\textsuperscript{\rm 1},
    Yi Yang\textsuperscript{\rm 1},
    Minfeng Zhu\textsuperscript{\rm 2}, \\
    Kecheng Zheng\textsuperscript{\rm 2},
    Shi Liu\textsuperscript{\rm 1},
    Wei Chen\textsuperscript{\rm 1}
}
\affiliations {
    \textsuperscript{\rm 1}State Key Lab of CAD\&CG, Zhejiang University\\
    \textsuperscript{\rm 2}Zhejiang University\\
}

\usepackage{bibentry}

\begin{document}

\maketitle

\begin{abstract}
Recent advancements in pre-trained Vision-Language Models (VLMs) have highlighted the significant potential of prompt tuning for adapting these models to a wide range of downstream tasks. 
However, existing prompt tuning methods typically map an image to a single representation, limiting the model's ability to capture the diverse ways an image can be described.
To address this limitation, we investigate the impact of visual prompts on the model’s generalization capability and introduce a novel method termed Multi-Representation Guided Prompt Tuning (MePT).
Specifically, MePT employs a three-branch framework that focuses on diverse salient regions, uncovering the inherent knowledge within images which is crucial for robust generalization.
Further, we employ efficient self-ensemble techniques to integrate these versatile image representations, allowing MePT to learn all conditional, marginal, and fine-grained distributions effectively.
We validate the effectiveness of MePT through extensive experiments, demonstrating significant improvements on both base-to-novel class prediction and domain generalization tasks.
\end{abstract}

%

\section{Introduction}
Pre-trained vision-language models (VLMs), such as CLIP \cite{CLIP}, have demonstrated remarkable performance across a wide range of downstream tasks in a zero-shot manner without task-specific fine-tuning.
For instance, an image can be classified by using CLIP to measure the similarities with multiple textual prompts (e.g. “A photo of a \{class\}.”).
It has been shown that the quality of text templates significantly impacts the performance of CLIP \cite{menon2023visual, CLIP}.
To this end, prompt tuning techniques have been developed, which not only alleviate the challenge of manually crafting prompts but also enhance the model’s generalization capabilities with minimal training data, while keeping the VLM model parameters frozen \cite{coop, cocoop, shu2022testPrompt, huang2022unsupervisedPrompt}.
However, a notable limitation of prompt tuning is its tendency to overfit task-specific data distributions, potentially leading to the forgetting of valuable knowledge acquired during the extensive large-scale pretraining phase \cite{cocoop}.
\begin{figure}[h]
    \centering
    \hspace{-0.035\textwidth}
    \includegraphics[width=0.5\textwidth]{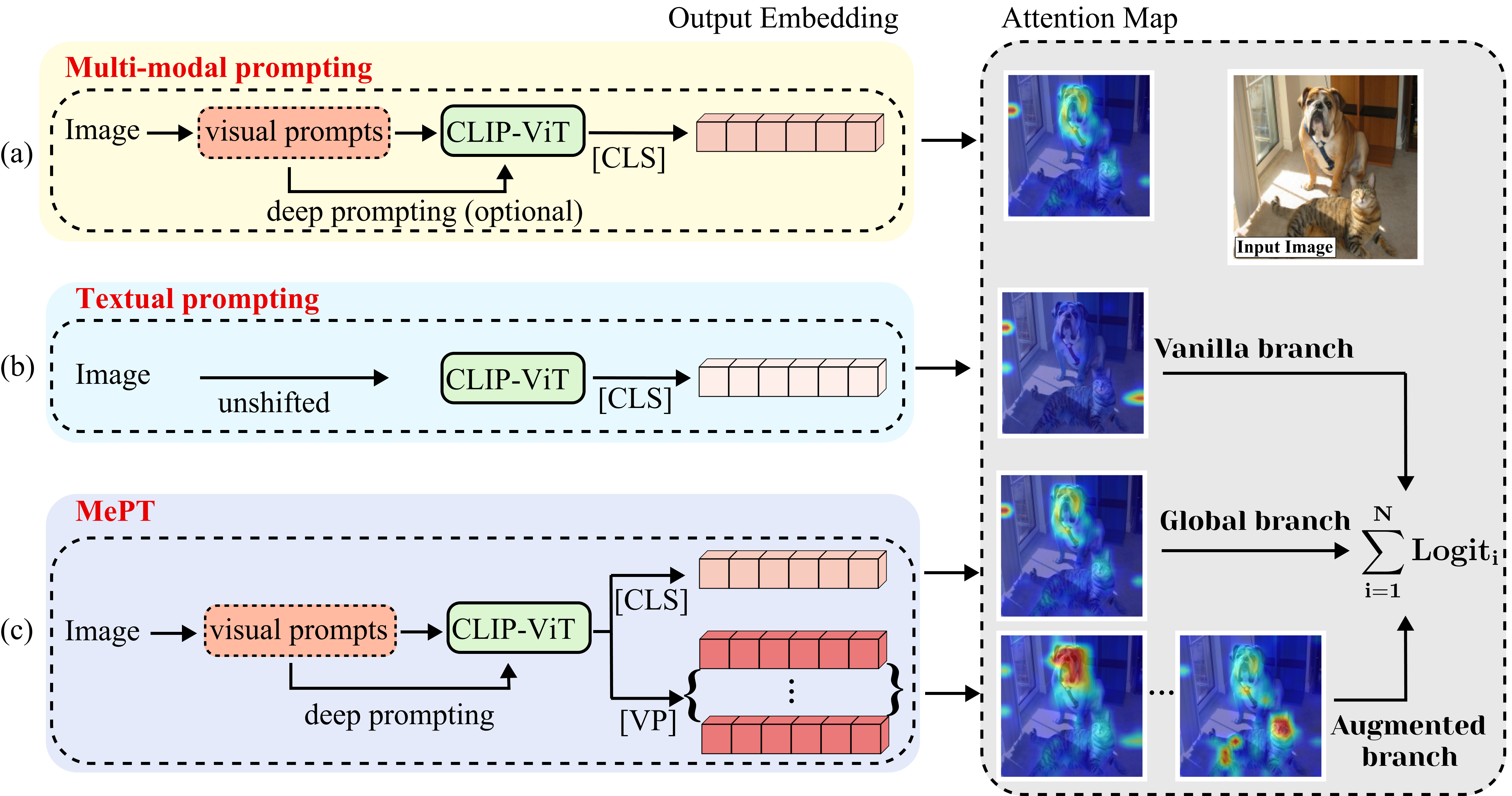}
    \caption{The illustration of multi-representation prompting. 
    Given an image, we visualize the attention maps from the last layer of the vision transformer (Right).
    After visual-side prompt tuning (a), CLIP effectively optimizes the relevance signal for foreground objects compared to vanilla CLIP (b).
    The visual prompts tokens ([VP]s) further demonstrate foreground-focused capabilities and naturally attend to different objects in the scene with diversity (c).
    We propose a novel Multi-Representation Guided Prompt Tuning framework, designed to capture the comprehensive information inherent in the image with three branches.
    %
    }
    \label{fig:intro_image_812}
\end{figure}

To mitigate the issue of overfitting, existing works concentrate on reducing the discrepancy between the text prompts and the general textual knowledge within pre-trained models. 
It has been approached with various strategies, including alignment with text templates \cite{kgcoop, tcp, lasp}, integration of extra text representations \cite{promptSRC, argue, CPL, chen2023plot} and gradient-based optimization methods \cite{ProGrad}.
In addition to these approaches, visual prompt tuning technology has been introduced, to enhance synergy between the vision and language representations \cite{vpt, maple}. 
However, existing prompt tuning techniques mainly rely on global image representations (i.e., the embedding derived from the special class token).
These approaches treat the image as one single point, ignoring the fact that different text prompts may only focus on one or a subset of visual characteristics \cite{chen2023plot, guo2023calip}.
As illustrated in Figure \ref{fig:intro_image_812}, relying solely on a global representation derived from the class token proves inadequate for capturing all the visual information embedded in images.
This limitation constrains both the effectiveness of image recognition and the robustness of domain generalization \cite{chefer_Optimizing, paiss2022no, geirhos2020shortcut}.

In efforts to incorporate more visual information, approaches such as PromptSRC \cite{promptSRC} and RPO \cite{RPO}, despite the adoption of visual prompts, emphasize the importance of primitive image representations generated by vanilla CLIP, particularly in the category and domain shift tasks.
PromptSRC aims to regulate prompted image representations by leveraging the foundational vanilla representations, while RPO utilizes multiple inner-masked visual prompts to address internal representation shifts.
In contrast, our method constructs a more comprehensive representation space and leverages the synergy of diverse image representations to mitigate overfitting.

Our method employs a three-branch framework, with each branch meticulously designed to capture role-specific visual representations.
The first branch termed the global branch, is pivotal as it aligns with the foundational design of the CLIP model, which is trained to match global visual features with corresponding language features.
This global image representation remains indispensable for a broad range of tasks, ensuring that the foundational knowledge is retained.
The second branch, referred to as the augmented branch, addresses the diverse attention patterns exhibited by learnable visual prompts.
As illustrated in Figure~\ref{fig:intro_image_812}, this branch utilizes outputs from visual prompts as augmented image representations, enabling a focus on different salient regions and tailoring the approach to domain-specific tasks.
Additionally, these visual prompts also optimize the foreground signal for global representations, thereby improving overall model performance, with further analysis provided in our experiments.
Finally, the vanilla branch is designed to preserve general visual knowledge, which is particularly beneficial for achieving generalization in unseen domains. 
To integrate the visual information captured by these three branches, we introduce a parameter-efficient self-ensemble strategy.
Similarly to the model ensemble technique \cite{kang2020novel, modelEN2022patching, modelEN2022robust}, the final prediction is derived from the aggregation of predictions made by each branch. 
This strategy enhances the model's robustness to domain shifts and ensures comprehensive utilization of diverse visual features.

In summary, the main contributions are as follows:
\begin{itemize}
    \item
    We note that the visual prompts not only enhance the model's generalization capability but also optimize the foreground signal for global representations.
    Simultaneously, visual prompts themselves reveal versatile attention patterns, naturally focusing on diverse salient regions of the attention map.
    \item 
    To enhance synergistic comprehension for images, we propose Multi-Representation Guided Prompt Tuning, a three-branch approach that effectively enhances robustness and generalization by ensembling global, augmented, and vanilla representations, ensuring comprehensive visual understanding.
    \item We conduct extensive experiments across 11 diverse datasets, collectively validating the effectiveness and robustness of our method on category shift and domain shift under few-shot generalization settings.
\end{itemize}

\begin{figure*}[ht]
    \centering
    \includegraphics[height=7.3cm, keepaspectratio]{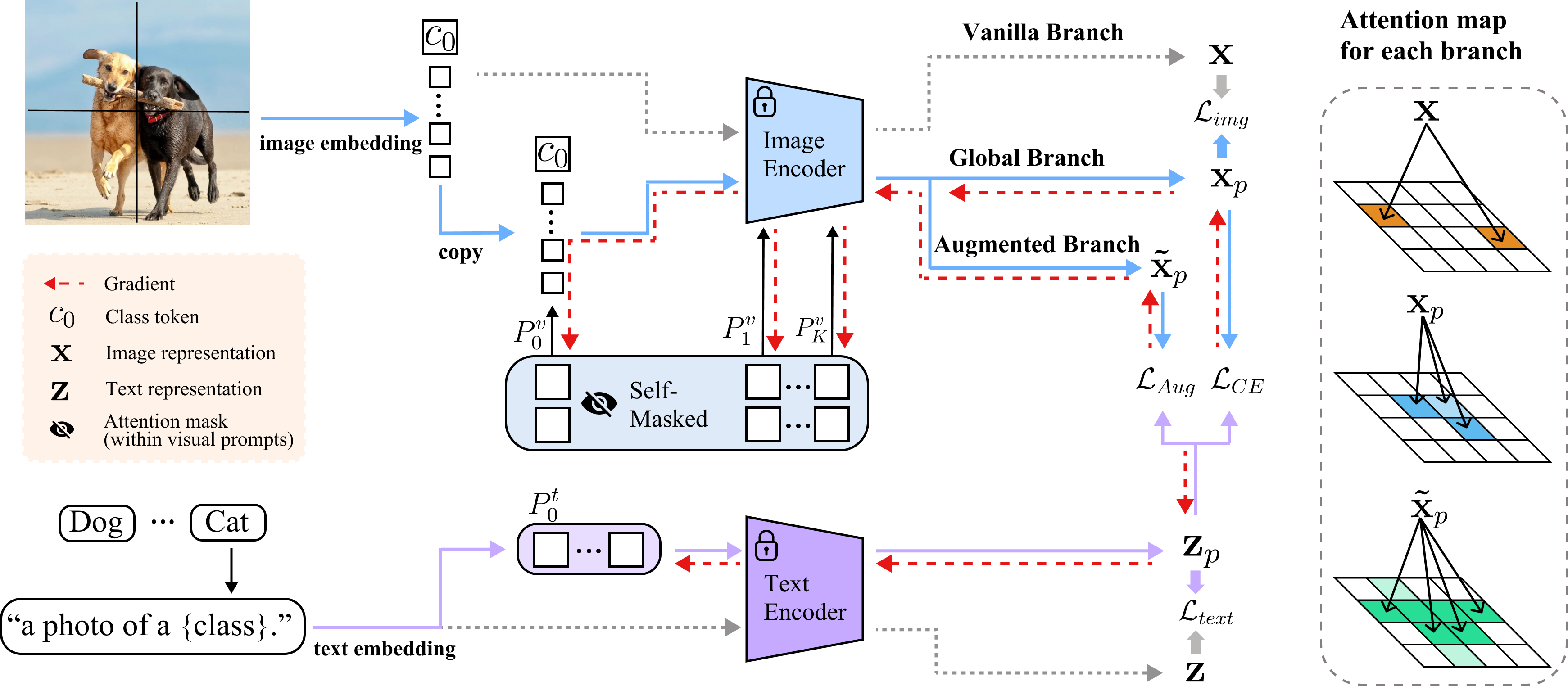}
    \caption{
    Overview of our proposed MePT framework for multi-modal prompt tuning.
    CLIP encoders are utilized to generate image and text representations from the input image-text pairs.
    We introduce three-branch image presentations: global representation $\mathbf{x_p}$, augmented representation $\mathbf{\tilde{x}_p}$, and vanilla representation $\mathbf{x}$ to ensure comprehensive visual understanding across different domains.
    Additionally, we employ ground truth supervision ($\mathcal{L}_{Aug}$) in the augmented branch and constraints ($\mathcal{L}_{img}$ and $\mathcal{L}_{text}$) in the global branch to align with general knowledge.
    Moreover, the self-masked attention within visual prompts is employed to restrict the attention flow. 
    }
    \label{fig:pipeline}
\end{figure*}

\section{Related Work}

\subsection{Prompt Tuning in Vision-Language models}
Prompt tuning, initially developed for Natural Language Processing (NLP), allows models to adapt efficiently to downstream tasks without the need to retrain previously learned parameters \cite{autoprompt:emnlp20, liu2021p, li2021prefix, lester2021power}.
In recent years, this technique has gained prominence in fine-tuning pre-trained VLMs, such as CLIP \cite{CLIP}. 
It involves introducing learnable text and visual prompts while keeping the pre-trained weights frozen \cite{maple, promptSRC, DAPT, ProDa}.
CoOp \cite{coop} first introduces prompt tuning into open-world visual understanding, enabling the adaptation of knowledge from large-scale vision-language pre-trained models.
Building on CoOp, its extended version, CoCoOp \cite{cocoop}, enhances the generalization of learnable textual prompts for each image.
To ensure that learnable text prompts retain essential general textual knowledge, approaches such as KgCoOp \cite{kgcoop}, ProGrad \cite{ProGrad}, and TCP \cite{tcp} have been developed. 
These methods constrain the prompts to align with foundational textual knowledge.
Additionally, visual prompt tuning \cite{vpt, yoo2023improving, oh2023blackvip} has been introduced to enhance the synergy between vision and language representations. 
MaPLe \cite{maple}, RPO \cite{RPO}, DAPT \cite{DAPT}, and promptSRC \cite{promptSRC} focus on tuning both vision and language branches in order to maintain cross-modal synergy.

\subsection{Visual Representation for Image Recognition}
Traditionally, CLIP and prompt tuning techniques have predominantly relied on global image representations generated by a special class token \cite{CLIP}. 
These tokens act as feature aggregators, providing a global summary of the input image.
Recent studies \cite{InterCLIP, reg, vpt, chen2023plot, chowdhury2023patch, sun2024alpha} have revealed that significant potential remains for capturing and utilizing the rich information embedded within images.
For instance, \cite{TEn} leverages multiple pre-trained or fine-tuned VLMs to generate customized ensembles of image and text representations for decision-making.
CLIP-Adapter \cite{gao2024clip} introduces an adapter mechanism to adjust visual or text embeddings, aiming to improve representation alignment. 
More closely related to our work, PLOT \cite{chen2023plot} extracts local visual features with varied semantic cues from feature maps. 
This method generates discriminative and visually-aligned local textual prompt tokens.
However, PLOT has also demonstrated that depending solely on these feature maps without alignment with textual prompts can lead to a significant drop in performance.
These studies underscore that relying solely on global representations derived from the class tokens is insufficient to capture the full range of information present in images.
%
%

\section{Method}
\subsection{Preliminary}
\paragraph{Overview.} 
In this section, we provide an overview of the multi-modal representations in CLIP related to prompt tuning. 
CLIP aims to explore the semantic correspondence between the vision and language modalities through large-scale pre-training, utilizing two encoders, a transformer-based text encoder $M_{text}$ and an image encoder $M_{img}$.

\paragraph{CLIP image and text representation.} 
The image $I$ and text description $t$ are first divided into $m$ patches and $n$ word tokens, which are then projected into patch embeddings $E_0 \in \mathbb{R}^{m \times d_v}$ and word embeddings $W_0 \in \mathbb{R}^{n \times d_t}$.
Additional special tokens, namely the \textit{class token} $c_0$ and \textit{eos token} $e_0$, are included and later used as the output tokens.
These embeddings $E_i$ and $W_i$ are inputted into the $(i + 1)$th layer of the corresponding encoder, as shown below:
\begin{equation}
\begin{aligned}
\left[ c_i, E_i, P_i^v \right] &= M_{img}^i \left( \left[ c_{i-1}, E_{i-1}, P_{i-1}^v \right] \right), \\
\vspace{0.5cm}
\left[ e_i, W_i, P_i^t \right] &= M_{text}^i \left( \left[ e_{i-1}, W_{i-1}, P_{i-1}^t \right] \right)
\end{aligned}
\label{eq:encoder}
\end{equation}
Here, $P_i^v$ $\in \mathbb{R}^{V \times d_v}$ and $P_i^t$ $\in \mathbb{R}^{T \times d_t}$, denote learnable prompt tokens appended after the patch and word embeddings respectively.
$V$ and $T$ are the number of learnable visual prompts and text prompts.
To obtain the final image representation $\mathbf{x}_p$ and text representation $\mathbf{z}_p$, the output embeddings of special tokens $c_K$ and $e_K$ after $K$-layer encoders, are projected to a common vision-language latent space:
\begin{equation}
\mathbf{x}_p = \mathbf{ImgProj}(c_K), \quad \mathbf{z}_p = \mathbf{TextProj}(e_K)
\label{eq:proj}
\end{equation}

\paragraph{Prompt Tuning for CLIP.} 
Prompt tuning, or multi-modal prompt tuning approaches, involve appending learnable prompts to either the image encoder or text encoder after original embeddings \cite{coop}, as described in Equation~\eqref{eq:encoder}.
Each vector has the same dimension as the original word or patch embedding. 
For image recognition on a target dataset $\mathcal{D}$, learnable prompts are optimized with the cross-entropy loss $\mathcal{L}_{CE}$:
\begin{equation}
\mathcal{L}_{CE} = -\sum_{\mathcal{D}} \log \frac{\exp \left( \text{sim}(\mathbf{x}_p, \mathbf{z}_p^y) / \tau \right)}
{\sum_{i=1}^{N_c} \exp \left( \text{sim}(\mathbf{x}_p, \mathbf{z}_p^i) / \tau \right)}
\label{eq:LCE}
\end{equation}
Here, $y$ represents the ground truth label for the input image.
By leveraging a few labeled samples, the prompts are fine-tuned specifically for the downstream task.

\subsection{Multi-Representation branches}
As illustrated in Figure~\ref{fig:pipeline}, we propose a three-branch approach, with each branch dedicated to capturing role-specific visual representations.
This allows MePT to learn all conditional, marginal, and fine-grained distributions effectively and uncovers the inherent knowledge within images that is essential for robust generalization.

\paragraph{Global branch.}
Since the original CLIP model is trained by aligning global visual features with language features, the global image representation is crucial for a wide range of tasks. 
To extend this capability, we introduce deep learnable visual prompts  $P^v = \{ P_0^v, P_1^v, \cdots, P_{K-1}^v \}$.
%
For a given input sample, we obtain the final output embedding $c_K$ by inserting learnable prompts at each layer:
\begin{equation}
\left[ c_i, E_i, \tilde{P}_i^v \right] = M_{img}^i \left( \left[ c_{i-1}, E_{i-1}, P_{i-1}^v \right] \right)
\label{eq:deep encoder}
\end{equation}
Here, $\tilde{P}_i^v$ refers to the output embedding of previous prompts $P_{i-1}^v$ after $i$-th layer image encoder, while in deep prompt tuning framework, it will be replaced soon by human-introduced prompts $P_i^v$.
The final prompted image representation is derived using the pre-trained projection matrix, as described in Equation~\eqref{eq:proj}, after the last vision transformer block.
This output of the global branch, denoted as $\mathbf{x}_p$, is referred to as the \textit{global representation}, acting as a global feature aggregator.

\paragraph{Augmented branch.}
Existing prompt-tuning methods map an image to a single global representation, which limits the model's ability to capture the diverse ways an image can be described.
To uncover the hidden knowledge of a single representation, we design the augmented branch, which enriches the image representations and exploits the foreground-focused ability of visual prompts.
Specially, we obtain the final visual prompts embeddings $\tilde{P}_K^{v}$ through the image encoder $M_{img}$, following Equation~\eqref{eq:deep encoder}.
These embeddings, typically discarded in previous prompt tuning methods, are then projected into the vision-language shared space to derive the corresponding representations $\mathbf{\tilde{x}}_p$:
\begin{equation}
\mathbf{\tilde{x}}_p = \mathbf{ImgProj}(\tilde{P}_K^v)
\label{eq:proj_vp}
\end{equation}
We refer $\mathbf{\tilde{x}}_p$ as \textit{augmented representation}.
These representations, associated with visual prompts, are able to focus on diverse salient regions of the attention map, as shown in Figure~\ref{fig:intro_image_812}, revealing excellent feature aggregation, which we will make more analysis in the next sections.

While visual prompts naturally develop the ability to focus on diverse salient regions during training, we find it beneficial to integrate a masking strategy within the visual prompts.
The visual self-masked attention mechanism ensures that additional visual prompts do not interfere with each other and restricts the flow of attention to other learnable prompts \cite{li2023blip}.
Unlike the read-only prompts used in RPO \cite{RPO}, our approach with the self-masked attention module imposes no additional restrictions on image patches, thereby preserving the flow of global attention.
The ablation study of the self-masked attention strategy will be provided in the next section.

\paragraph{Vanilla branch.}
Even though CLIP's image and text encoder weights are kept frozen, learnable prompts can influence the internal representations within the self-attention module \cite{reg, touvron2021training, RPO}, potentially affecting performance variance and generalization, particularly in data-deficient settings.
To this end, we utilize the vanilla image representation $\mathbf{x}$ without extra learnable visual prompts appended, therefore preserving the integrity of the original features.
We refer to $\mathbf{x}$ as \textit{vanilla representation}.
This vanilla branch preserves general visual knowledge and proves to be beneficial for generalization to unseen categories and domains.

\subsection{Prompt Tuning with multi-representation}
\paragraph{Global strategy.}
Following the prompt tuning process, we minimize the cross-entropy loss $\mathcal{L}_{CE}$, as described in Equation~\eqref{eq:LCE}.
However, strong downstream dataset transfer has been shown to cause prompts to overfit task-specific data, limiting their ability to effectively utilize the general information from the frozen model \cite{kgcoop, promptSRC}.
To address this, we impose a constraint on the prompted visual and text representations to ensure their consistency with the vanilla CLIP representations by using cosine similarity:
\begin{equation}
\mathcal{L}_{text} = 1-\text{sim} (\mathbf{x}_p, \mathbf{x}), \quad
\mathcal{L}_{img} = 1-\text{sim}(\mathbf{z}_p, \mathbf{z})
\end{equation}
In summary, we aggregate the introduced components:
\begin{equation}
\mathcal{L}_{Global} = \mathcal{L}_{CE} + \lambda_1 \mathcal{L}_{text} + \lambda_2 \mathcal{L}_{img}
\end{equation}
%
The $\mathcal{L}_{Global}$ loss guides the global branch to acquire complementary knowledge from both the downstream tasks and the pre-trained CLIP model, with $\lambda_1$ and $\lambda_2$ as the corresponding hyperparameters.

\paragraph{Augmented strategy.}
Visual prompts, as previously discussed, exhibit versatile attention patterns that enable them to focus on various objects within a scene.
Following Equation~\eqref{eq:proj_vp}, by introducing $V$ individual visual prompt tokens, we obtain augmented representations $\mathbf{\tilde{x}}_p = \{ \mathbf{\tilde{x}}_p^1, \mathbf{\tilde{x}}_p^2, \cdots, \mathbf{\tilde{x}}_p^V \}$.
These representations are then used to calculate similarity scores with the corresponding text representations.
The final logits are obtained by averaging the similarity scores with the equal weight:
\begin{equation}
\text{sim}(\mathbf{\tilde{x}}_p, \mathbf{z}_p) = \frac{1}{V} \sum_{i=1}^{V} \frac{\mathbf{\tilde{x}}_p^i \cdot \mathbf{z}_p}{\|\mathbf{\tilde{x}}_p^i\| \|\mathbf{z}_p\|}
\end{equation}
%
Although the augmented branch is designed to capture diverse salient regions, it requires additional supervision to fully leverage its potential for downstream image recognition tasks, particularly when starting from random initialization \cite{wang2024revisiting}. 
To harness the full potential of these visual prompts, we propose an augmented strategy tailored specifically for domain-specific image representations:
\begin{equation}
\mathcal{L}_{Aug} =
-\sum_{\mathcal{D}} \log \frac{\exp \left( \text{sim}(\mathbf{\tilde{x}}_p, \mathbf{z}_p^y) / \tau \right)}
{\sum_{i=1}^{N_c} \exp \left( \text{sim}(\mathbf{\tilde{x}}_p, \mathbf{z}_p^i) / \tau \right)}
\end{equation}
Here, we incorporate classification supervision using ground truth labels to enhance the branch's few-shot visual recognition. $N_c$ is the number of seen classes.
Consequently, our overall training objective is defined as:
\begin{equation}
\mathcal{L} = \mathcal{L}_{Global} + \mathcal{L}_{Aug}
\end{equation}

\paragraph{Representations with self-ensemble.}
Unlike methods that map an image to a single representation, our MePT method proposes a three-branch framework for fine-grained comprehension of each image.
%
We employ a parameter-efficient self-ensemble strategy that integrates global image representation $\mathbf{x}_p$, domain-augmented representation $\mathbf{\tilde{x}}_p$, and primitive image representation $\mathbf{x}$ to enhance robustness to domain shifts.
By computing the similarity between each image representation and the corresponding text representations, we derived predictions for each branch: $p(y \vert \mathbf{x}_p)$, $p(y \vert \mathbf{\tilde{x}}_p)$, $p(y \vert \mathbf{x})$. 
These outputs are then combined using our logit-level self-ensemble strategy, yielding the final classification results.
Notably, unlike the traditional ensemble technique \cite{kang2020novel, TEn}, our self-ensemble strategy is implemented within a single-phase prompt tuning process and does not require a secondary model for conditioning.
A comparative analysis of various ensemble strategies, such as confidence weighting \cite{TEn} and thresholding method \cite{bai2024prompt}, will be discussed in our ablation experiments.

\section{Experiment}

\subsection{Benchmark Setting}

\paragraph{Base-to-novel class generalization.}
We assess the generalizability of MePT by adopting a benchmark setting where the datasets are divided into base and novel classes, as outlined in \cite{cocoop}. 
The model is initially trained solely on the base classes in a few-shot setting and is then evaluated on both the base and novel classes to measure its ability to generalize across different categories.

\paragraph{Cross-dataset generalization.}
To assess the robustness of our approach in transferring to unseen datasets, we directly evaluate our ImageNet-trained model on a variety of previously unseen datasets, without performing any data-specific fine-tuning. 
This allows us to validate the model's ability to generalize effectively across different domains.

\paragraph{Quantitative segmentation.}
We follow a standard protocol for evaluating heatmap-based methods \cite{chefer2021generic, chefer2021transformer} to validate the proposed method. 
In this process, the raw attention map is considered a foreground segmentation of the image and is compared against the ground truth segmentation of the dataset. 
The performance of the method is quantitatively assessed using three key metrics: (i) pixel accuracy (pixAcc), (ii) mean Intersection over Union (mIoU), and (iii) mean Average Precision (mAP).

\paragraph{Datasets.}
To evaluate our method's generalization capability from base to novel classes and its performance across diverse datasets, we conduct experiments on 11 image recognition datasets: ImageNet \cite{imagenet}, Caltech101 \cite{caltech}, OxfordPets \cite{oxfordpet}, StanfordCars \cite{stanfordcars}, Flowers102 \cite{flower}, Food101 \cite{food}, FGVCAircraft \cite{fgvcair}, SUN397 \cite{sun}, DTD \cite{dtd}, EuroSAT \cite{helber2019eurosat}, and UCF101 \cite{ucf101}.
To ensure a fair comparison, we adhere to the protocol established by CoCoOp \cite{cocoop}, randomly sampling 16 images per class for the training set. 
Each experiment is repeated three times with different random seeds, and the average performance is reported.
Additionally, we assess segmentation quality on ImageNet-Segmentation \cite{imagenet-seg}, a subset of the ImageNet validation set comprising 4,276 images with annotated segmentations.


\begin{table}[!t]
\resizebox{\columnwidth}{!}{%
\setcellgapes{3.25pt}
\makegapedcells
\begin{tabular}{cccccccc}
\Xhline{3\arrayrulewidth}
Dataset                     & \multicolumn{1}{c|}{}     & CoOp  & CoCoOp & RPO            & MaPLe          & \multicolumn{1}{c|}{\begin{tabular}[c]{@{}c@{}}\makecell{Prompt-\\SRC}\end{tabular}} & \begin{tabular}[c]{@{}c@{}}\makecell{MePT\\(Ours)}\end{tabular}          \\ \hline
\multirow{3}{*}{Average}    & \multicolumn{1}{c|}{Base} & 82.69 & 80.47  & 81.13          & 82.28          & \multicolumn{1}{c|}{84.26}                                                & \textbf{84.63} \\
                            & \multicolumn{1}{c|}{Novel}  & 63.22 & 71.69  & 75.00          & 75.14          & \multicolumn{1}{c|}{76.10}                                                & \textbf{76.30} \\
                            & \multicolumn{1}{c|}{HM}    & 71.66 & 75.83  & 77.78          & 78.55          & \multicolumn{1}{c|}{79.97}                                                & \textbf{80.25} \\ \hline
\multirow{3}{*}{ImageNet}   & \multicolumn{1}{c|}{Base} & 76.47 & 75.98  & 76.60          & 76.66          & \multicolumn{1}{c|}{77.60}                                               & \textbf{77.96} \\
                            & \multicolumn{1}{c|}{Novel}  & 67.88 & 70.43  & \textbf{71.57} & 70.54          & \multicolumn{1}{c|}{70.73}                                                & 69.88          \\
                            & \multicolumn{1}{c|}{HM}    & 71.92 & 73.10  & 74.00          & 73.47          & \multicolumn{1}{c|}{\textbf{74.01}}                                       & 73.70          \\ \hline
\multirow{3}{*}{Caltech101} & \multicolumn{1}{c|}{Base} & 98.00 & 97.96  & 97.97          & 97.74          & \multicolumn{1}{c|}{98.10}                                                & \textbf{98.47} \\
                            & \multicolumn{1}{c|}{Novel}  & 89.81 & 93.81  & 94.37          & 94.36          & \multicolumn{1}{c|}{94.03}                                                & \textbf{94.39} \\
                            & \multicolumn{1}{c|}{HM}    & 93.73 & 95.84  & 96.03          & 96.02          & \multicolumn{1}{c|}{96.02}                                                & \textbf{96.39} \\ \hline
\multirow{3}{*}{OxfordPets} & \multicolumn{1}{c|}{Base} & 93.67 & 95.20  & 94.63          & 95.43          & \multicolumn{1}{c|}{95.33}                                                & \textbf{96.09} \\
                            & \multicolumn{1}{c|}{Novel}  & 95.29 & 97.69  & 97.50          & \textbf{97.76}          & \multicolumn{1}{c|}{97.30}                                                & 97.45 \\
                            & \multicolumn{1}{c|}{HM}    & 94.47 & 96.43  & 96.05          & 96.58          & \multicolumn{1}{c|}{96.30}                                                & \textbf{96.77} \\ \hline
\multirow{3}{*}{Cars}       & \multicolumn{1}{c|}{Base} & 78.12 & 70.49  & 73.87          & 72.94          & \multicolumn{1}{c|}{78.27}                                                & \textbf{79.93} \\
                            & \multicolumn{1}{c|}{Novel}  & 60.40 & 73.59  & \textbf{75.53}          & 74.00          & \multicolumn{1}{c|}{74.97}                                       & 73.54          \\
                            & \multicolumn{1}{c|}{HM}    & 68.13 & 72.01  & 74.69          & 73.47          & \multicolumn{1}{c|}{76.58}                                                & \textbf{76.60} \\ \hline
\multirow{3}{*}{Flowers}    & \multicolumn{1}{c|}{Base} & 97.60 & 94.87  & 94.13          & 95.92          & \multicolumn{1}{c|}{\textbf{98.07}}                                       & 97.94          \\
                            & \multicolumn{1}{c|}{Novel}  & 59.67 & 71.75  & \textbf{76.67} & 72.46          & \multicolumn{1}{c|}{76.50}                                                & 76.15          \\
                            & \multicolumn{1}{c|}{HM}    & 74.06 & 81.71  & 84.50          & 82.56          & \multicolumn{1}{c|}{\textbf{85.95}}                                       & 85.68          \\ \hline
\multirow{3}{*}{Food101}    & \multicolumn{1}{c|}{Base} & 88.33 & 90.70  & 90.33          & \textbf{90.71} & \multicolumn{1}{c|}{90.67}                                                & 90.58          \\
                            & \multicolumn{1}{c|}{Novel}  & 82.26 & 91.29  & 90.83          & \textbf{92.05} & \multicolumn{1}{c|}{91.53}                                                & 91.41          \\
                            & \multicolumn{1}{c|}{HM}    & 85.19 & 90.99  & 90.58          & \textbf{91.38} & \multicolumn{1}{c|}{91.10}                                                & 90.99          \\ \hline
\multirow{3}{*}{Aircraft}   & \multicolumn{1}{c|}{Base} & 40.44 & 33.41  & 37.33          & 37.44          & \multicolumn{1}{c|}{42.73}                                                & \textbf{43.74} \\
                            & \multicolumn{1}{c|}{Novel}  & 22.30 & 23.71  & 34.20          & 35.61          & \multicolumn{1}{c|}{\textbf{37.87}}                                       & 36.79          \\
                            & \multicolumn{1}{c|}{HM}    & 28.75 & 27.74  & 35.70          & 36.50          & \multicolumn{1}{c|}{\textbf{40.15}}                                       & 39.97          \\ \hline
\multirow{3}{*}{SUN397}     & \multicolumn{1}{c|}{Base} & 80.60 & 79.74  & 80.60          & 80.82          & \multicolumn{1}{c|}{\textbf{82.67}}                                       & 82.53          \\
                            & \multicolumn{1}{c|}{Novel}  & 65.89 & 76.86  & 77.80          & \textbf{78.70}          & \multicolumn{1}{c|}{78.47}                                       & 77.43          \\
                            & \multicolumn{1}{c|}{HM}    & 72.51 & 78.27  & 79.18          & 79.75          & \multicolumn{1}{c|}{\textbf{80.52}}                                       & 79.90          \\ \hline
\multirow{3}{*}{DTD}        & \multicolumn{1}{c|}{Base} & 79.44 & 77.01  & 76.70          & 80.36          & \multicolumn{1}{c|}{\textbf{83.37}}                                       & 82.64          \\
                            & \multicolumn{1}{c|}{Novel}  & 41.18 & 56.00  & 62.13          & 59.18          & \multicolumn{1}{c|}{62.97}                                                & \textbf{66.22} \\
                            & \multicolumn{1}{c|}{HM}    & 54.24 & 64.85  & 68.61          & 68.16          & \multicolumn{1}{c|}{71.75}                                                & \textbf{73.52} \\ \hline
\multirow{3}{*}{EuroSAT}    & \multicolumn{1}{c|}{Base} & 92.19 & 87.49  & 86.63          & \textbf{94.07} & \multicolumn{1}{c|}{92.90}                                                & 94.00          \\
                            & \multicolumn{1}{c|}{Novel}  & 54.74 & 60.04  & 68.97          & 73.23          & \multicolumn{1}{c|}{73.90}                                                & \textbf{76.85} \\
                            & \multicolumn{1}{c|}{HM}    & 68.69 & 71.21  & 76.79          & 82.35          & \multicolumn{1}{c|}{82.32}                                                & \textbf{84.56} \\ \hline
\multirow{3}{*}{UCF101}     & \multicolumn{1}{c|}{Base} & 84.69 & 82.33  & 83.67          & 83.00          & \multicolumn{1}{c|}{\textbf{87.10}}                                                & 87.04          \\
                            & \multicolumn{1}{c|}{Novel}  & 56.05 & 73.45  & 75.43          & 78.66          & \multicolumn{1}{c|}{78.80}                                                & \textbf{79.14} \\
                            & \multicolumn{1}{c|}{HM}    & 67.46 & 77.64  & 79.34          & 80.77          & \multicolumn{1}{c|}{82.74}                                                & \textbf{82.90} \\ \Xhline{3\arrayrulewidth}            
\end{tabular}%
}
\caption{Comparison with state-of-the-art methods on base-to-novel class prediction. 
HM represents the harmonic mean of the test accuracy on base and novel classes.
} 
\label{tab: base2novel}
\end{table}

\begin{table*}[!t]
\resizebox{\textwidth}{!}{%
\setlength{\tabcolsep}{5pt} 
\renewcommand{\arraystretch}{1.3} 
\begin{tabular}{ccccccccccccc}
\Xhline{2.5\arrayrulewidth}
          & Source         & \multicolumn{11}{c}{Target}                                                                                                                                                              \\ \cmidrule(lr){2-2}   \cmidrule{3-13}
          & ImageNet       & Caltech101     & OxfordPets     & Cars           & Flowers        & Food101        & Aircraft       & SUN397         & DTD            & EuroSAT        & UCF101         & Average        \\ \hline
CLIP      & 66.72          & 92.94          & 89.07          & 65.29          & 71.30          & 86.11          & \textbf{24.87} & 62.62          & 44.56          & 47.69          & 66.77          & 65.12          \\
CoOp      & 71.51          & 93.70          & 89.14          & 64.51          & 68.71          & 85.30          & 18.47          & 64.15          & 41.92          & 46.39          & 66.55          & 63.88          \\
CoCoOp    & 71.02          & \textbf{94.43} & 90.14          & 65.32          & \textbf{71.88} & 86.06          & 22.94          & 67.36          & 45.73          & 45.37          & 68.21          & 65.74          \\
PromptSRC & 71.27          & 93.60          & 90.25          & \textbf{65.70} & 70.25          & \textbf{86.15} & 23.90          & 67.10          & \textbf{46.87} & 45.50          & 68.75          & 65.81          \\ \hline
\rowcolor[HTML]{EFEFEF} 
MePT      & \textbf{72.09} & 93.54          & \textbf{90.71} & 65.47          & 70.93          & 85.81          & 23.57          & \textbf{67.70} & 45.61          & \textbf{48.38} & \textbf{69.38} & \textbf{66.11} \\ \Xhline{2.5\arrayrulewidth}
\end{tabular}%
}
\caption{Comparison of MePT with existing advanced approaches on cross-dataset benchmark evaluation.} 
\label{tab: cross_dataset}
\end{table*}

\paragraph{Implementation details.}
In all experiments, we employ a pre-trained CLIP model with a ViT-B/16 vision encoder backbone \cite{dosovitskiy2021an} as our foundational architecture.
For the base-to-novel class generalization setting, we employ deep visual and text prompt tuning, with prompts initialized randomly using a normal distribution.
The learning rate is fixed at 0.0016, with a batch size of 32, and training is conducted over 50 epochs.
We assign $\lambda_1$ = 3 and $\lambda_2$ = 4 to appropriately weight the text loss $\mathcal{L}_{text}$ and image loss $\mathcal{L}_{img}$, respectively.
The length of the text prompts, $T$, is set to 4, while the length of the visual prompts, $V$, is set to 32.
For cross-dataset generalization, we set the visual prompt length $V$ to 8, with the learning rate fixed at 0.05 and the number of training epochs at 10.
In the segmentation evaluation, to more clearly demonstrate the impact of visual prompts on image representation, we exclusively employ deep visual prompt tuning \cite{vpt} without incorporating additional modules.
Refer to the Appendix for additional implementation details.

\subsection{Base-to-Novel Generalization}
We compare the performance of our approach with CoOp \cite{coop}, CoCoOp \cite{cocoop}, RPO \cite{RPO}, MaPLe \cite{maple}, and PromptSRC \cite{promptSRC}.
Table~\ref{tab: base2novel} provides a comparative analysis of our methods against these baseline models for base and novel class prediction, demonstrating that MePT consistently outperforms previous methods.
Specifically, MePT significantly enhances performance over CoCoOp by 4.16\% for base classes and 4.61\% for novel classes, and over the deep multi-modal prompt tuning method MaPLe by 2.35\% and 1.16\%, respectively.
We attribute these improvements to the integration of both vanilla and augmented representations, which effectively bridge the distributional gap between the downstream tasks and the pretraining dataset, thereby enhancing performance.
Notably, on datasets with a larger distributional gap from ImageNet \cite{maple}, such as EuroSAT (a satellite imagery dataset) and DTD (a texture dataset), MePT demonstrates substantial improvements. For instance, MePT achieves a 13\% performance gain over CoCoOp on the EuroSAT dataset.
The overall trend suggests that MePT's effectiveness increases with the diversity of the dataset. 
However, it is important to note that we do not employ class-specific context or text-side augmentation, which results in lower performance on fine-grained classification datasets. 
For instance, both CoOp and MePT, without text-side augmentation, underperform compared to PromptSRC on the Flower dataset.

\begin{table}[!t]
\resizebox{\columnwidth}{!}{%
\setlength{\tabcolsep}{6.5pt} 
\renewcommand{\arraystretch}{1.5} 
\begin{tabular}{lcc|c}
\Xhline{2.5\arrayrulewidth}
Method                           & \multicolumn{1}{c}{Base Acc.} & \multicolumn{1}{c|}{Novel Acc.} & \multicolumn{1}{c}{HM} \\ \hline
Independent V-L prompting (IVLP) & 83.65                         & 70.64                         & 76.60                  \\
1: Vanilla branch                   & 77.65                         & \textbf{75.30}                         & 76.46                  \\
2: Augmented branch                 & \underline{83.88}                         & 71.23                        & \underline{77.04}                  \\
3: Global branch                    & \textbf{84.03}                         & \underline{75.17}                         & \textbf{79.35}                  \\ \Xhline{2.5\arrayrulewidth}
\end{tabular}%
}
\caption{
Effect of our proposed three-branch framework. 
Results are presented for each individual branch.
Underlined values indicate sub-optimal performance.
}
\label{tab: each branch}
\end{table}

\subsection{Cross-Dataset Evaluation}
We evaluate the cross-dataset generalization capability of MePT by training multi-modal prompts on all 1000 ImageNet classes and then directly applying them to the remaining 10 datasets. 
Table~\ref{tab: cross_dataset} presents the cross-dataset evaluation results, comparing our method with state-of-the-art approaches.
Overall, our tuning method exhibits the best generalization performance across the target datasets while maintaining robust classification capabilities on the ImageNet source dataset. 
Compared to CoOp, our approach shows competitive results and achieves superior generalization on 9 out of 10 datasets.
%

\begin{table}[!t]
\resizebox{\columnwidth}{!}{%
\setlength{\tabcolsep}{5pt} 
\renewcommand{\arraystretch}{1.5} 
\begin{tabular}{lcc|c}
\Xhline{2.5\arrayrulewidth}
Method                              & Base Acc. & Novel Acc. & HM    \\ \hline
1: Global                           & 84.03     & 75.17    & 79.35 \\
2: Global + Vanilla               & 83.30     & \underline{76.04}    & 79.50 \\
3: Global + Vanilla + Augment     & \underline{84.57}     & 75.86    & \underline{79.98} \\
\rowcolor[HTML]{EFEFEF} 
4: Global + Promitive + maskAgument & \textbf{84.63}     & \textbf{76.30}    & \textbf{80.25} \\ \Xhline{2.5\arrayrulewidth}
\end{tabular}%
}
\caption{
Effect of our proposed multi-representation techniques. 
Results are averaged over 11 datasets.} 
\label{tab: global+}
\end{table}

\begin{table}[!t]
\resizebox{\columnwidth}{!}{%
\setlength{\tabcolsep}{7pt} 
\renewcommand{\arraystretch}{1.5} 
\begin{tabular}{lcc|c}
\Xhline{2.5\arrayrulewidth}
Method                           & \multicolumn{1}{c}{Base Acc.} & \multicolumn{1}{c|}{Novel Acc.} & \multicolumn{1}{c}{HM} \\ \hline
1: Confidence weighting          & 84.04                         & \textbf{76.30}                         & 79.98                  \\
2: Threshold (\textgreater 0.8)     & 84.55                         & 76.18                         & 80.15                  \\
\rowcolor[HTML]{EFEFEF} 
3: Equal weighting (mean)                      & \textbf{84.63}                         & \textbf{76.30}                         & \textbf{80.25}                  \\ \Xhline{2.5\arrayrulewidth}
\end{tabular}%
}
\caption{
Comparison of branch ensemble strategies. 
Equal-weighted branch ensemble strategy demonstrates superior performance over others.
} 
\label{tab: ensemble strategy}
\end{table}

\begin{figure*}[!h]
    \centering
    \includegraphics[width=0.95\textwidth, height=4cm]{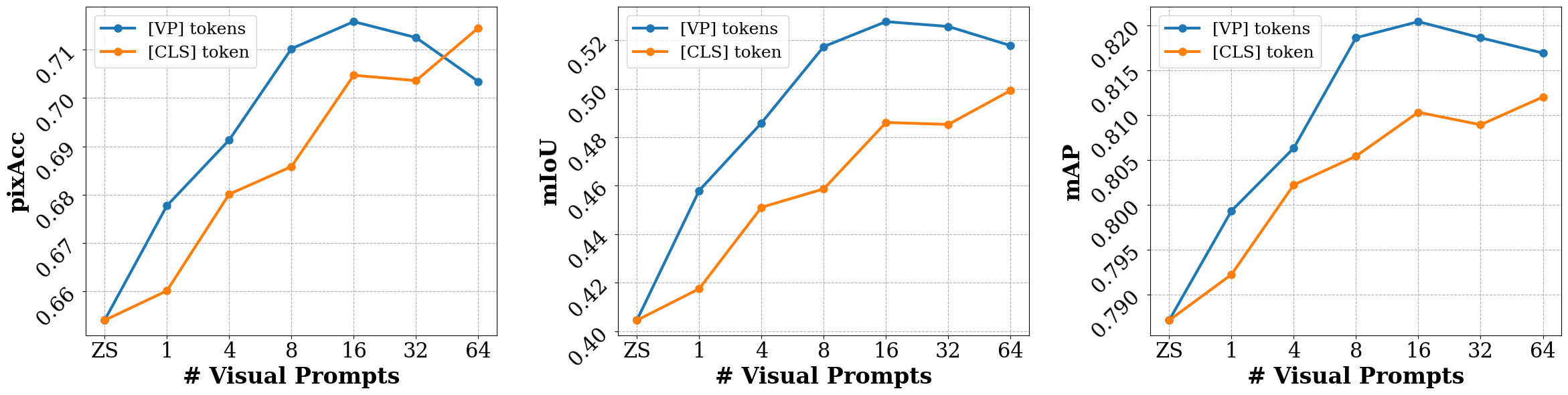}
    \caption{
    Comparison of foreground/background segmentation results of the special class token ([CLS]) and visual prompts tokens ([VP]s) with zero-shot CLIP baseline.
    The visual prompts are trained on ImageNet with 16-shot per base class.
    }
    \label{fig: seg_results}
\end{figure*}

\subsection{Segmentation Results}
As shown in Figure~\ref{fig: seg_results}, with the evaluation of attention-based explainability methods, we analyze the raw attention map outputs of the images from both the class token and visual prompts.
These outputs are binarized to generate foreground/background segmentation maps.
By comparing the segmentation quality with the zero-shot results produced by the vanilla CLIP, we observe that, after few-shot visual prompts tuning, the segmentation accuracy consistently surpassed that of zero-shot methods across all metrics.
The improvements are closely correlated with the number of visual prompt tokens utilized, suggesting that visual prompt tuning effectively optimizes the relevance signal for foreground objects while reducing the model's reliance on irrelevant background cues.
Further, the visual prompts exhibit foreground representational capabilities that are comparable to, and in some cases exceed, those of the special class token. 
These visual prompts naturally attend to diverse objects of the attention map, even when initialized randomly.

\subsection{Ablation Experiments}
\paragraph{Effectiveness of three-branch presentations.}
We employ a simple baseline method, Independent Vision-Language Prompting (IVLP) \cite{maple}, which learns deep prompt tokens separately on both the text and image encoders. 
To evaluate the individual contributions of our multi-representation branches, we disentangle them within our prompting framework, as shown in Table~\ref{tab: each branch} and Table~\ref{tab: global+}.
%
Incorporating the vanilla branch results in a significant 4.66\% increase in novel class performance, although it leads to a 6.00\% decrease in base class performance.
This indicates that the vanilla branch effectively enforces prompts to capture generalizable features from the frozen CLIP model, yet struggles with domain-specific tasks.
For the augmented branch, performance improvements are observed in both base and novel classes, with gains of 0.23\% and 0.59\%, respectively.
Additionally, in this branch, both base class and novel class performance surpass the IVLP baseline, achieving near-optimal results. 
This suggests that, with ground truth supervision, the visual prompts are capable of effectively classifying and recognizing both in-domain and out-of-domain distributions.
As for the global branch, it leads to significant performance gains, with base class accuracy increasing by 0.39\% and a substantial improvement of 4.53\% in novel class accuracy.
%
%

%
Moreover, as illustrated in Table~\ref{tab: global+}, by ensembling these role-specific branches, we note that the vanilla branch substantially enhances performance on unseen categories, while the self-masked augmented branch predominantly improves performance on seen categories.

\paragraph{Comparison on ensemble strategies.}
Table~\ref{tab: ensemble strategy} presents a comparative study of different prompt ensemble techniques.
We evaluate our approach against two baseline methods.
The first baseline utilizes a confidence-aware weighting ensemble \cite{TEn, Liu_2023_ICCV}, which adjusts the logits of different branches according to their respective confidence levels.
The second baseline utilizes a logit level thresholding method \cite{bai2024prompt, pmlr-v139-feng21f}, applying a threshold of 0.8 to the outputs of the three branches.
In contrast, our equal weighting aggregation strategy achieves the highest average performance across both base and novel classes across 11 datasets.

\begin{figure}[t]
    \centering
    \hspace{-0.03\textwidth}
    \begin{minipage}[b]{0.235\textwidth}
        \centering
        \includegraphics[width=\textwidth]{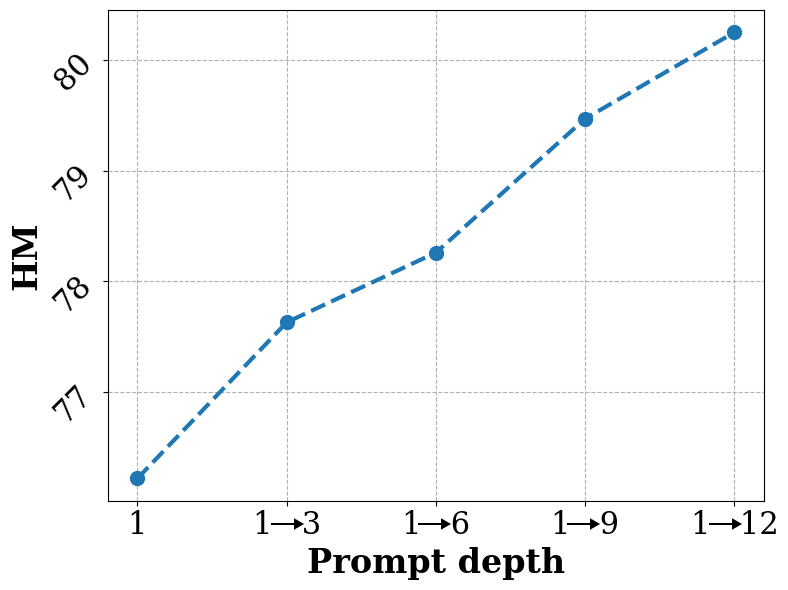}
        \label{fig: prompt depth}
    \end{minipage}
    \begin{minipage}[b]{0.235\textwidth}
        \centering
        \includegraphics[width=\textwidth]{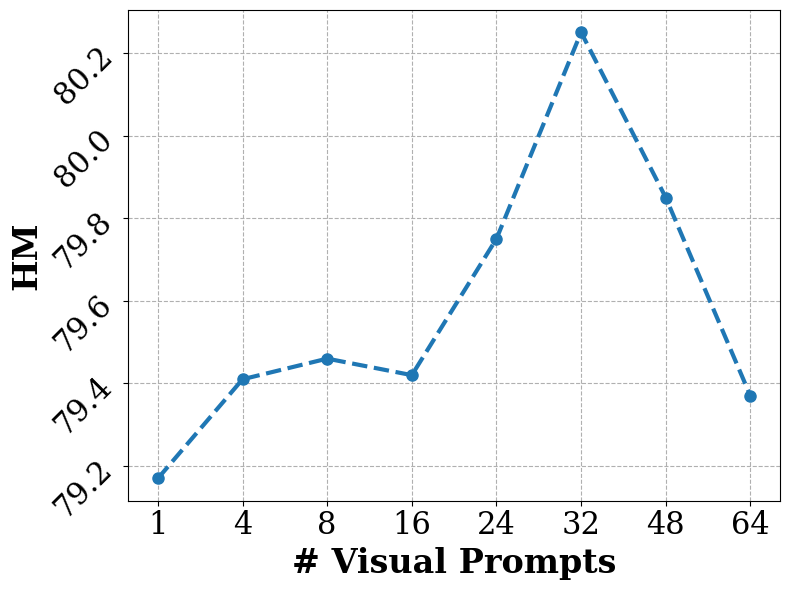}
        \label{fig: prompt length}
    \end{minipage}
    \caption{Ablation study on prompt depth (left) and visual prompt length (right) over 11 datasets.}
    \label{fig: ablation_depth_length}
\end{figure}

\paragraph{Prompt depth.}
Figure~\ref{fig: ablation_depth_length} (left) illustrates the impact of prompt depth on performance. 
Generally, performance improves as the prompt depth increases. 
This trend is consistent with findings reported by \cite{vpt, yoo2023improving}, where employing deep vision-language prompts across all layers yields the highest harmonic mean.

\paragraph{Prompt length.}
Figure~\ref{fig: ablation_depth_length} (right) presents the effect of visual prompt length on the harmonic mean. 
Overall, performance initially improves as the prompt length increases, but after reaching an optimal point, it begins to decline. 
Specifically, the highest harmonic mean is achieved with a prompt length of 32 vision prompts.

%
%


\section{Conclusion}
In this paper, we propose Multi-Representation Guided Prompt Tuning (MePT), a novel approach designed to enhance robustness and generalization in vision-language models. 
Our method employs a three-branch strategy that ensembles global, augmented, and vanilla image representations, thereby providing a synergistic and comprehensive understanding of visual information across a wide range of domains.
The experimental results demonstrate that our approach significantly improves performance on image recognition tasks across various datasets, validating its effectiveness and robustness.
%
Our method provides a direction for enriching image representations through visual prompts, but further research is needed to fully explore the impact of these visual prompts and the underlying mechanisms through which they affect the overall image representation.

%

\newpage

\bibliography{aaai25}

\newpage\clearpage

\section{Supplementary Material}

The supplementary material contents are organized in the following order:
\begin{itemize}
    \item Additional implementation details (Appendix A)
    
    \item Domain generalization experiments (Appendix B)  
    
    \item GradCAM segmentation results (Appendix C)

    \item Additional loss ablation study (Appendix D)

    \item Qualitative results of attention map (Appendix E)
    
\end{itemize}

\section{A. Additional implementation details}
\paragraph{Additional training details.}
In all experiments, we employ a pre-trained CLIP model with a ViT-B/16 vision encoder backbone \cite{dosovitskiy2021an}.
All models are trained using the SGD optimizer and utilize NVIDIA RTX 4090D GPU.
In segmentation evaluation experiments, to more clearly demonstrate the impact of visual prompts on image representation, we exclusively employ deep visual prompt tuning \cite{vpt} without incorporating additional modules, and the learning rate is fixed at 0.002.
We train on base ImageNet \cite{imagenet} classes with 16-shot to be consistent with the base-to-novel experiments.

\paragraph{Domain generalization settings.}
Additionally, we evaluate the robustness of our method on out-of-distribution datasets.
The domain generalization capability of MePT is assessed by training multi-modal prompts on all 1000 ImageNet classes and then directly applying them to four other ImageNet variants that contain various types of domain shifts.
The four ImageNet variants include ImageNetV2 \cite{imagenetv2},  ImageNet-Sketch \cite{sket}, ImageNet-A \cite{imagenetA} and ImageNet-R \cite{imagenetR}.
In these domain generalization experiments, all hyperparameters are kept the same as in the cross-dataset settings. 
We employ deep visual and text prompt tuning, with prompts initialized randomly using a normal distribution.
The learning rate is fixed at 0.05, with a batch size of 32, and training is conducted over 10 epochs.
We assign $\lambda_1$ = 3 and $\lambda_2$ = 4 to appropriately weight the text loss $\mathcal{L}_{text}$ and image loss $\mathcal{L}_{img}$, respectively.
The visual prompt length $V$ is set to 8 and the text prompt length $T$ is set to 4.

\section{B. Domain generalization}

Table \ref{tab: Domain generalization} presents the domain generalization results, comparing our method with state-of-the-art approaches. 
Overall, our tuning method exhibits favorable generalization performance across the four ImageNet variants while maintaining robust classification capabilities on the ImageNet source dataset. 
Specifically, MePT significantly improves performance on ImageNetV2 by 0.92\% compared to the previous state-of-the-art method, promptSRC.
Compared to CoCoOp and MaPLe, our approach achieves superior generalization on 3 out of 4 datasets and shows excellent overall results.

\begin{table}[!h]
\resizebox{\columnwidth}{!}{%
\setlength{\tabcolsep}{7pt} 
\renewcommand{\arraystretch}{1.3} 
\begin{tabular}{ccccccc}
\Xhline{2.5\arrayrulewidth}
          & Source         & \multicolumn{5}{c}{Target}                     \\ \cmidrule(lr){2-2}   \cmidrule{3-7} 
          & ImageNet       & -V2            & -S    & -A    & -R    & Avg.  \\ \hline
CLIP      & 66.73          & 60.83          & 46.15 & 47.77 & 73.96 & 57.18 \\
CoOp      & 71.51          & 64.20          & 47.99 & 49.71 & 75.21 & 59.28 \\
CoCoOp    & 71.02          & 64.07          & 48.75 & 50.63 & 76.18 & 59.91 \\
MaPLe     & 70.72          & 64.07          & 49.15 & \textbf{50.90} & 76.98 & 60.27 \\
PromptSRC & 71.27          & 64.35          & \textbf{49.55} & 50.90 & \textbf{77.80} & \textbf{60.65} \\ \hline
\rowcolor[HTML]{EFEFEF} 
MePT      & \textbf{72.09} & \textbf{65.27} & 49.39 & 49.56 & 77.21 & 60.36 \\ \Xhline{2.5\arrayrulewidth}
\end{tabular}%
}
\caption{Comparison of MePT with existing advanced approaches on domain generalization benchmark evaluation.} 
\label{tab: Domain generalization}
\end{table}

\begin{figure*}[!t]
    \centering
    \includegraphics[width=0.95\textwidth, height=4cm]{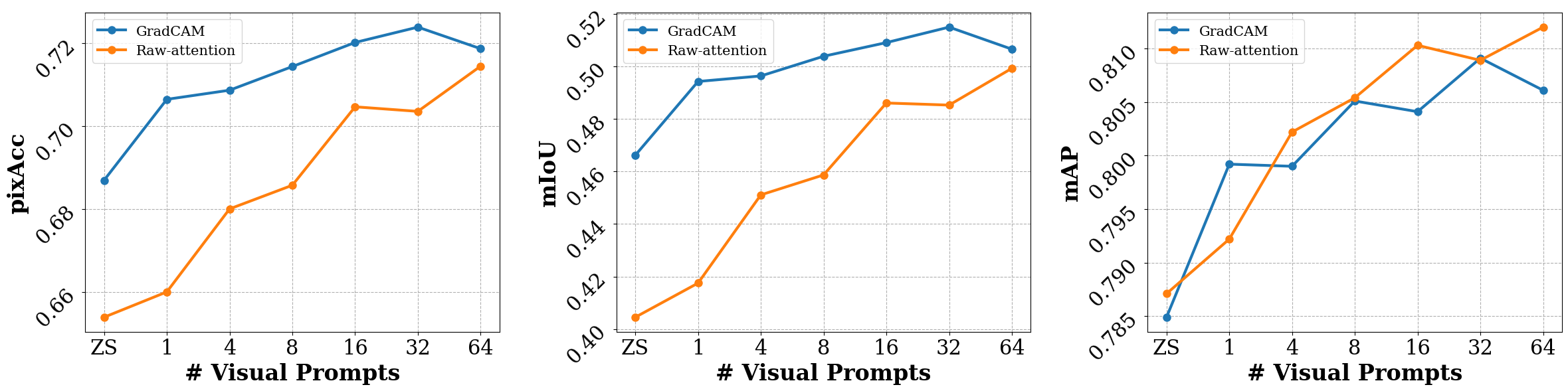}
    \caption{
    Comparison of foreground/background segmentation results between the GradCAM and raw attention map using the output embeddings derived from the special class token ([CLS]), relative to the zero-shot CLIP baseline.
    The visual prompts are trained on ImageNet with 16-shot per base class.
    }
    \label{fig: sup_seg_results}
\end{figure*}

\section{C. GradCAM segmentation results}
We also evaluate segmentation performance with GradCAM \cite{selvaraju2017grad}.
GradCAM is a class-specific visualization technique that combines the input features with the gradients of the network’s final layer to highlight salient regions in the input image.
The GradCAM segmentation evaluation settings are identical to those used for the raw attention map analysis mentioned earlier.
For GradCAM, we use the textual category that obtains the highest similarity score with the image to compute the gradient that is propagated.
As illustrated in Figure \ref{fig: sup_seg_results}, with few-shot visual prompt tuning and using the output embeddings derived from the special class token ([CLS]), both the class-agnostic behavior of the raw attention map and the class-specific behavior of GradCAM demonstrate significant improvement in highlighting foreground objects.

\section{D. Additional loss ablation study}
Table \ref{tab: loss ablation} presents ablation studies of losses employed in the MePT framework, namely $\mathcal{L}_{Aug}$, $\mathcal{L}_{img}$ and $\mathcal{L}_{text}$. 
The loss $\mathcal{L}_{Aug}$ provides additional supervision, fully leveraging the potential of visual prompts for downstream image recognition tasks and domain-specific image representation.
The ablation study shows that removing $\mathcal{L}_{Aug}$ results in a significant performance drop of 2.46\% on base classes, with only a minimal impact on novel classes, causing the 0.13\% drop.
This indicates that $\mathcal{L}_{Aug}$ is crucial to improve performance on base classes.
Conversely, $\mathcal{L}_{img}$ and $\mathcal{L}_{text}$ are employed to constrain the prompted visual and text representations, ensuring consistency with the vanilla CLIP representation.
The ablation experiment results show that removing $\mathcal{L}_{img}$ and $\mathcal{L}_{text}$ leads to a continuous performance drop of 2.93\% on novel classes, while there is a performance gain of 0.61\% on base classes.
This suggests that $\mathcal{L}_{img}$ and $\mathcal{L}_{text}$ are crucial for maintaining performance on novel classes, but their presence may limit the model's ability to effectively recognize images in seen classes.

\begin{table}[!h]
\resizebox{\columnwidth}{!}{%
\setlength{\tabcolsep}{9pt} 
\renewcommand{\arraystretch}{1.5} 
\begin{tabular}{lll|l}
\Xhline{2.5\arrayrulewidth}
Method                                   & \multicolumn{1}{c}{Base Acc.} & \multicolumn{1}{c|}{Novel Acc.} & \multicolumn{1}{c}{HM} \\ \hline
w/o $\mathcal{L}_{Aug}$, $\mathcal{L}_{text}$, $\mathcal{L}_{img}$ & 82.78                         & 73.24                         & 77.72                  \\ 
w/o $\mathcal{L}_{Aug}$                           & 82.17                         & 76.17                         & 79.06                  \\
\rowcolor[HTML]{EFEFEF} 
MePT                                     & 84.63                         & 76.30                         & 80.25 \\
\Xhline{2.5\arrayrulewidth}
\end{tabular}%
}
\caption{Effect of losses employed in MePT framework.} 
\label{tab: loss ablation}
\end{table}

\section{E. Qualitative results of attention map}
\begin{figure*}[]
    \centering
    \includegraphics[width=1.0\textwidth]{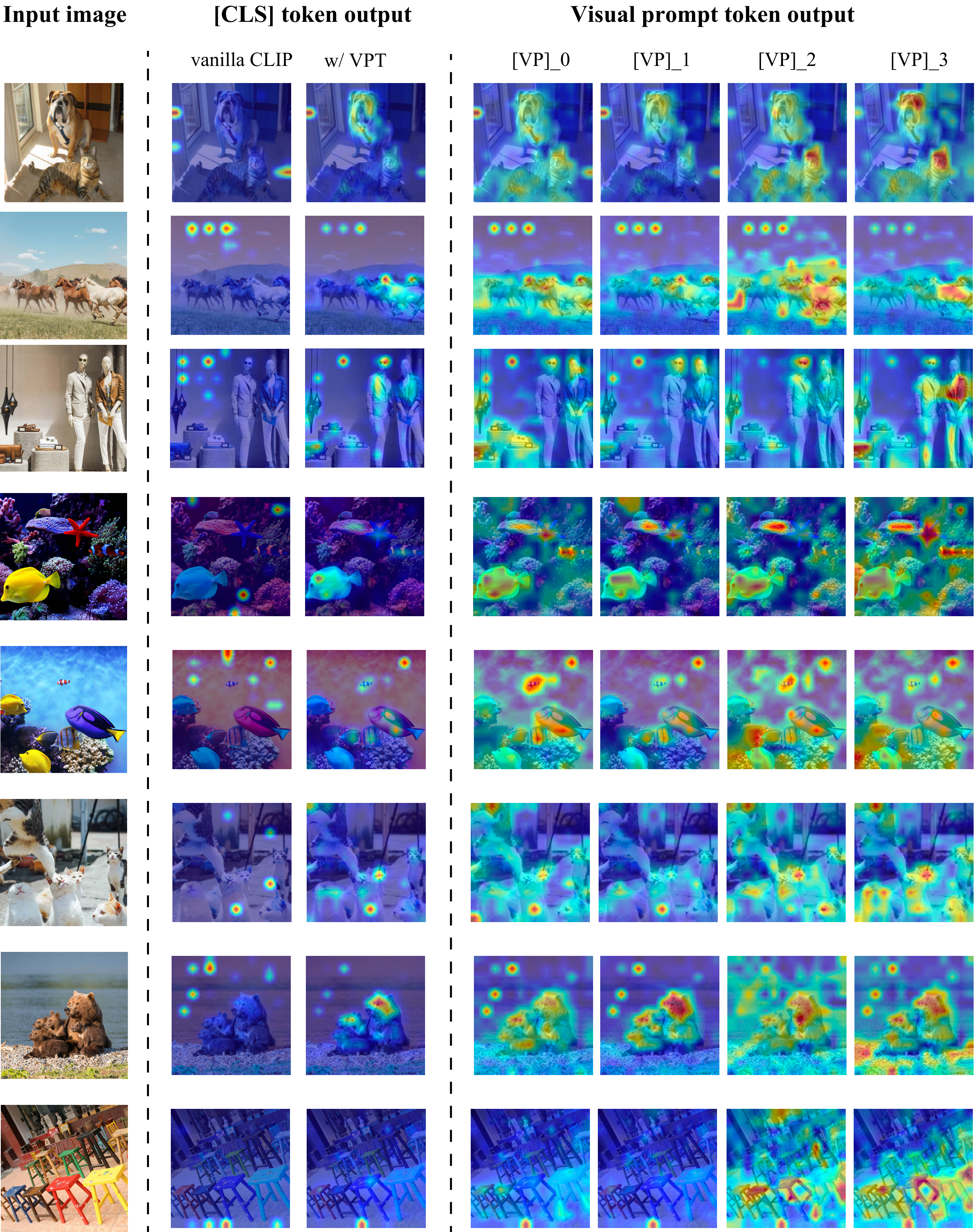}
    \caption{
    Attention maps of the class token ([CLS]) and the first four visual prompt tokens ([VP]s) to the patch tokens.
    The visual prompts are trained on ImageNet with 16-shot per category.
    }
    \label{fig: sup_visualize_results}
\end{figure*}

In Figure \ref{fig: sup_visualize_results}, we visualize the attention maps from the last layer of the Vision Transformer, using visual prompts trained on ImageNet with 16 shots per category.
Our approach generates diverse attention maps that highlight more salient regions on foreground objects while reducing reliance on irrelevant background cues. 
This enhanced focus on foreground areas explains the observed improvements in downstream image recognition tasks, particularly in category and domain shift scenarios under few-shot generalization settings.

\clearpage

\newpage\clearpage

\end{document}